\documentclass[10pt,twocolumn,letterpaper]{article}

\usepackage[pagenumbers]{cvpr} % To force page numbers, e.g. for an arXiv version
\usepackage{graphicx}
\usepackage{amsmath}
\usepackage{amssymb}
\usepackage{booktabs}

\usepackage[ruled,vlined]{algorithm2e}

\usepackage[pagebackref,breaklinks,colorlinks]{hyperref}

\usepackage[capitalize]{cleveref}
\crefname{section}{Sec.}{Secs.}
\Crefname{section}{Section}{Sections}
\Crefname{table}{Table}{Tables}
\crefname{table}{Tab.}{Tabs.}

\begin{document}

%junxiao-2022/11/4
\title{\texttt{PMR}: \texttt{P}rototypical \texttt{M}odal \texttt{R}ebalance for Multimodal Learning}

%\title{PMR: An universal rebalance method for multimodal learning with prototypes}

\author{%
   Yunfeng~Fan$^1$,
   Wenchao~Xu$^1$,
   Haozhao~Wang$^2$,
   Junxiao~Wang$^1$,
   and~Song~Guo$^1$\\
  \textsuperscript{1}Department of Computing, The Hong Kong Polytechnic University\\
  \textsuperscript{2}School of Computer Science and Technology, Huazhong University of Science and Technology\\
  \texttt{yunfeng.fan@connect.polyu.hk}, \texttt{wenchao.xu@polyu.edu.hk}, \texttt{hz\_wang@hust.edu.cn}, \\
  \texttt{junxiao.wang@polyu.edu.hk}, \texttt{song.guo@polyu.edu.hk}
}
\maketitle

%%%%%%%%% ABSTRACT
\begin{abstract}

% 2022/11/10 fyf
Multimodal learning (MML) aims to jointly exploit the common priors of different modalities to compensate for their inherent limitations.
However, existing MML methods often optimize a uniform objective for different modalities, leading to the notorious ``modality imbalance'' problem and counterproductive MML performance. 
To address the problem, some existing methods modulate the learning pace based on the fused modality, which is dominated by the better modality and eventually results in a limited improvement on the worse modal. 
To better exploit the features of multimodal, we propose Prototypical Modality Rebalance (PMR) to perform stimulation on the particular slow-learning modality without interference from other modalities.  
Specifically, we introduce the prototypes that represent general features for each class, to build the non-parametric classifiers for uni-modal performance evaluation. Then, we try to accelerate the slow-learning modality by enhancing its clustering toward prototypes. 
Furthermore, to alleviate the suppression from the dominant modality, we introduce a prototype-based entropy regularization term during the early training stage to prevent premature convergence.
Besides, our method only relies on the representations of each modality and without restrictions from model structures and fusion methods, making it with great application potential for various scenarios.
%
% The source code is available at. 

\end{abstract}

%%%%%%%%% BODY TEXT
\section{Introduction}
\label{sec:intro}

%% multimodal learning has potential 
%% problem: model imbalance （reasons）
%% current solution defects
%% the reason why these solution has defects  (highlight main challenges)
%% we found something: (magnitude and its direction)
%% our method definition
%% why our method can solve the problem
%% benefits: performance, universal （强调一下支持universal的好处：不仅仅是模型通用性 已有通用性 未来通用性 ）

Multimodal learning (MML)~\cite{ngiam2011multimodal,srivastava2012multimodal,shankar2018review} emerges to mimic the way humans perceive the world, i.e., from multiple sense channels toward a common phenomenon for better understanding the external environment, which has attracted extensive attention in various scenarios, e.g. video classification~\cite{jiang2018modeling,tian2019multimodal,pandeya2021deep}, event localization~\cite{tian2018audio,xia2022cross}, action recognition~\cite{shahroudy2017deep,imran2020evaluating}, audiovisual speech recognition~\cite{mroueh2015deep,oneactua2022improving}. By employing a complementary manner of multimodal training, it is expected that MML can achieve better performance than using a single modality.  
However, the heterogeneity of multimodal data poses challenges on how to learn multimodal correlations and complementarities.

\begin{figure}[t]
    \centering
    \includegraphics[width=1.1\linewidth]{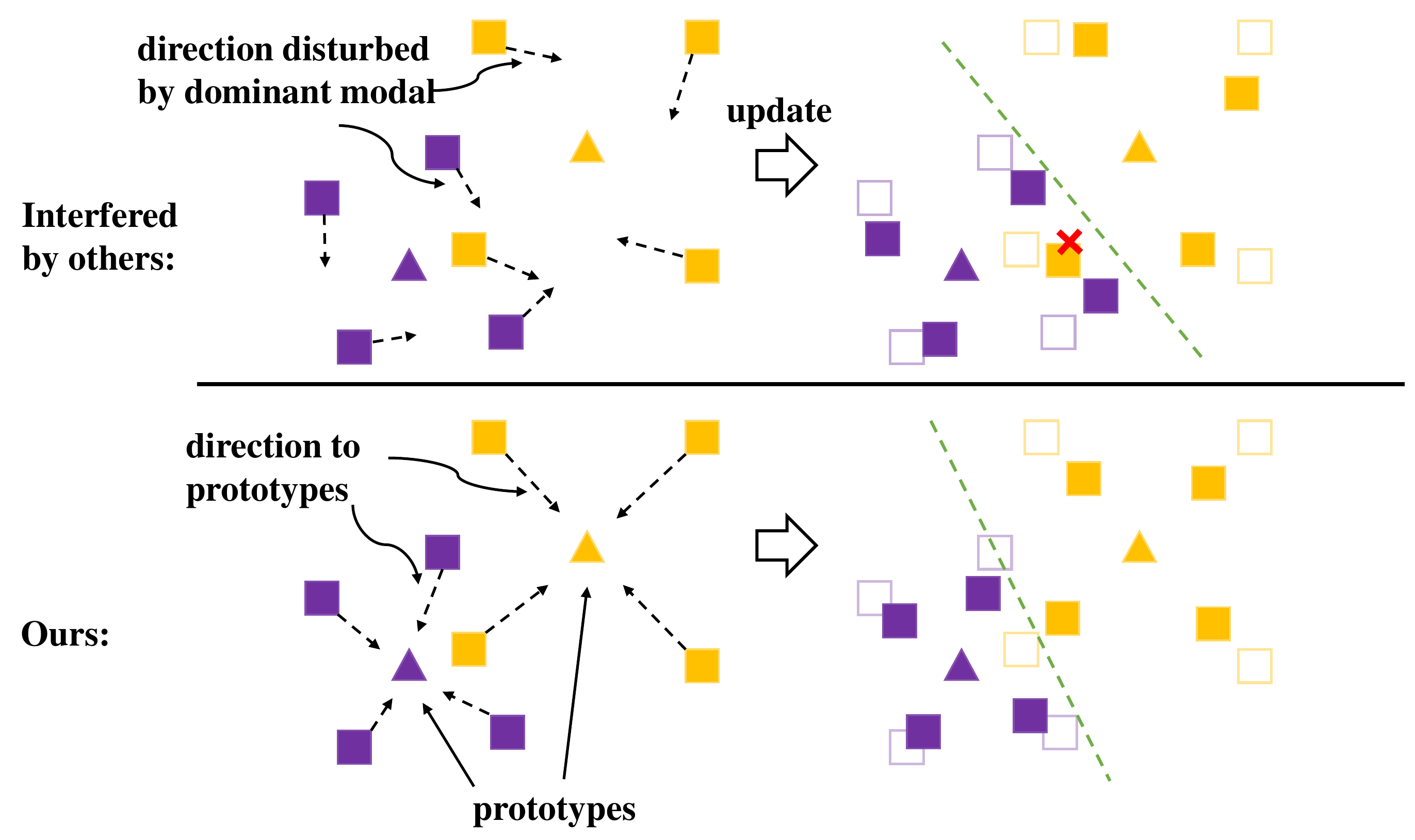}
    \caption{The slow-learning modal's updating direction is severely disturbed by the dominant one, making it hard to exploit its features. We propose to use the prototypes, the centroids of each class in representation space, to adjust updating direction for better uni-modal performance. Other modalities will not interfere with the new direction, which ensures improvement.}
    \label{fig:prior and ours}
\end{figure}

%junxiao-2022/11/4
According to recent research~\cite{peng2022balanced,wu2022characterizing}, although the overall performance of multimodal learning exceeds that of single-modal learning, the performance of each modality tends to be far from their upper bound.
The reason behind this phenomenon is the ``modality imbalance'' problem, in which the dominant modality will hinder the full utilization of multimodal.
The researchers~\cite{wang2020makes} also claimed that different modalities overfit and converge at different rates, meaning that optimizing the same objective for different modalities leads to inconsistent learning efficiency.

Several methods~\cite{wang2020makes,peng2022balanced,xiao2020audiovisual} have been proposed to address the problem.
Some of them~\cite{peng2022balanced,xiao2020audiovisual} try to modulate the learning paces of different modalities based on the fusion modal. 
However, We find out through experiments that the dominant modality not only suppresses the learning rates of other modalities~\cite{peng2022balanced} but also interferes with their update direction, which makes it hard to improve the performance of slow-learning modalities.
Moreover, existing methods either inevitably bring additional model structures~\cite{du2021improving,wang2020makes} or are limited by specific fusion methods~\cite{peng2022balanced,wu2022characterizing}, which limit their application scenarios.

To tackle their limitations, we propose the \textbf{P}rototypical \textbf{M}odal \textbf{R}ebalance (PMR) strategy to stimulate the slow-learning modality via promoting the exploitation of features and alleviate the suppression from the dominant modality by slowing down itself in the early training stage. 

Concretely, we introduce the prototypes for each modality, which are defined as ``representative embeddings of instances of a class''. We utilize the prototypes to construct non-parametric classifiers by comparing the distances between each sample with all the prototypes to evaluate the performance of each modality and design a new prototype-based metric inspired by \cite{peng2022balanced} to monitor the modality imbalance degree during the training process.  
Then, we propose the prototypical cross-entropy (PCE) loss to accelerate the slow-learning modality by enhancing its clustering process, as illustrated in \cref{fig:prior and ours}. 
The PCE loss can achieve comparable performance to the cross-entropy (CE) loss~\cite{arora2019theoretical} in the classification task, and more importantly, it is not affected by the dominant modality and gives internal impetus for full feature exploitation instead of going on in the disturbed direction.
In addition, we introduce a prototypical entropy regularization (PER) term, which can be seen as a penalty on the dominant modality to prevent premature convergence for suppression effect alleviation.
Our method only relies on the representations of each modality and without restrictions from model structures and fusion methods.
Therefore, the PMR strategy has great generality potential. 
To summarize, our contributions in this paper are as follows:

\begin{itemize}
\item We analyze the modal imbalance problem and find that during the training process, the deviation of the gradient update direction of the uni-modal became larger, indicating that we should not regulate along the original gradient.

\item We propose PMR to address the modal imbalance problem by actively accelerating slow-learning modalities with PCE loss and simultaneously alleviating the suppression of the dominant modality via PER. 

\item We conduct comprehensive experiments and demonstrate that 1) PMR can achieve considerable improvements over existing methods; 2) PMR is independent of the fusion method or model structure and has strong advantages in generality.
\end{itemize}

%-------------------------------------------------------------------------

\section{Related Works}
\label{sec: related}
\subsection{Imbalanced multimodal learning}
Several recent studies~\cite{wang2020makes,du2021improving,winterbottom2020modality} have shown that many multimodal DNNs cannot achieve better performance compared to the best single-modal DNNs.
Wang \etal~\cite{wang2020makes} found that different modalities overfit and generalize at different rates and thus obtain suboptimal solutions when jointly training them using a unified optimization strategy.
Peng \etal~\cite{peng2022balanced} proposed that the better-performing modality will dominate the gradient update while suppressing the learning process of the other modality.
Furthermore, it has been reported that multimodal DNNs can exploit modal bias in the data, which is inconsistent with the expectation of exploiting cross-modal interactions in VQA~\cite{jabri2016revisiting,goyal2017making,winterbottom2020modality}.

Several approaches have been proposed recently to deal with the modal imbalance problem~\cite{wang2020makes,xiao2020audiovisual,peng2022balanced,du2021improving}.
Wang \etal~\cite{wang2020makes} used additional classifiers for each modality and its fusion modality and then optimized the gradient mixing problem they introduced to obtain better weights for each branch.
Du \etal~\cite{du2021improving} tried to improve uni-modal performance by distilling knowledge from well-trained models.
However, these approaches introduce more model structure and computational effort, which makes the training process more complex and expensive.
Xiao \etal~\cite{xiao2020audiovisual} proposed DropPathway, which randomly drops the audio pathway during training as a regularization technique to adjust the learning paces between visual and audio pathways.
Peng \etal~\cite{peng2022balanced} chose to slow down the learning rate of the mighty modality by online modulation to lessen the inhibitory effect on the other modality.
Although a certain degree of improvement is achieved, such approaches do not impose the intrinsic motivation of improvement on the slow-learning modality, making the improvement of this modality a passive rather than an active behavior.
Besides, the interference from other modalities will hinder the improvement by modulation based on the fused modality data.
Furthermore, the application scenarios of these methods are limited by fusion methods or model structures.
In this paper, we aim to power the slow-learning modality to reduce the gap with the dominant one and allow it to be used in conjunction with various fusion methods and model architectures.

\begin{figure*}[t]
  \centering
  \begin{subfigure}{0.325\linewidth}
    %\fbox{\rule{0pt}{2in} \rule{.9\linewidth}{0pt}}
    \includegraphics[width=0.9\linewidth]{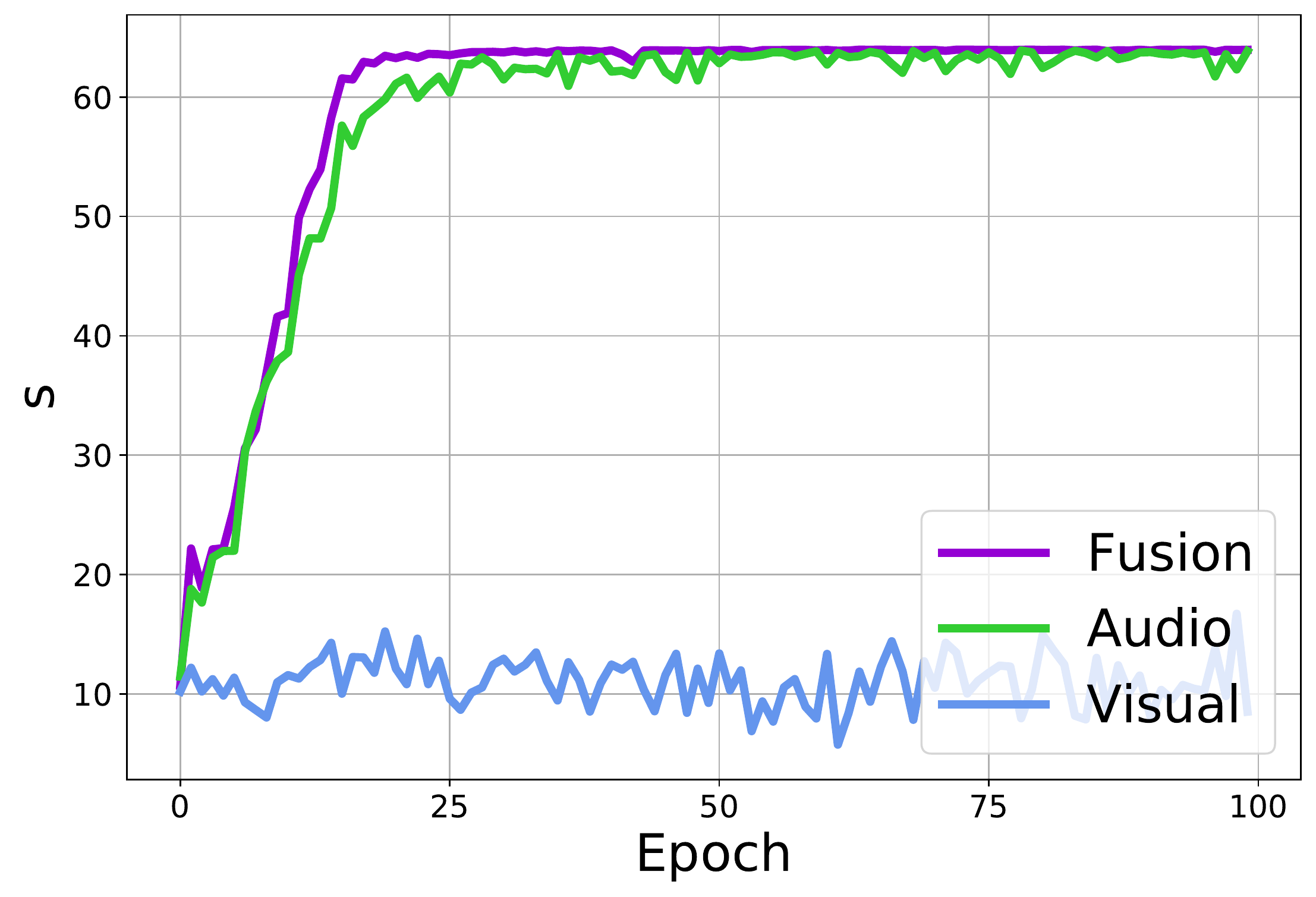}
    \caption{}
    \label{fig:performance comparison of each modality and their fusion}
  \end{subfigure}
  \hfill
  \begin{subfigure}{0.325\linewidth}
    %\fbox{\rule{0pt}{2in} \rule{.9\linewidth}{0pt}}
    \includegraphics[width=0.9\linewidth]{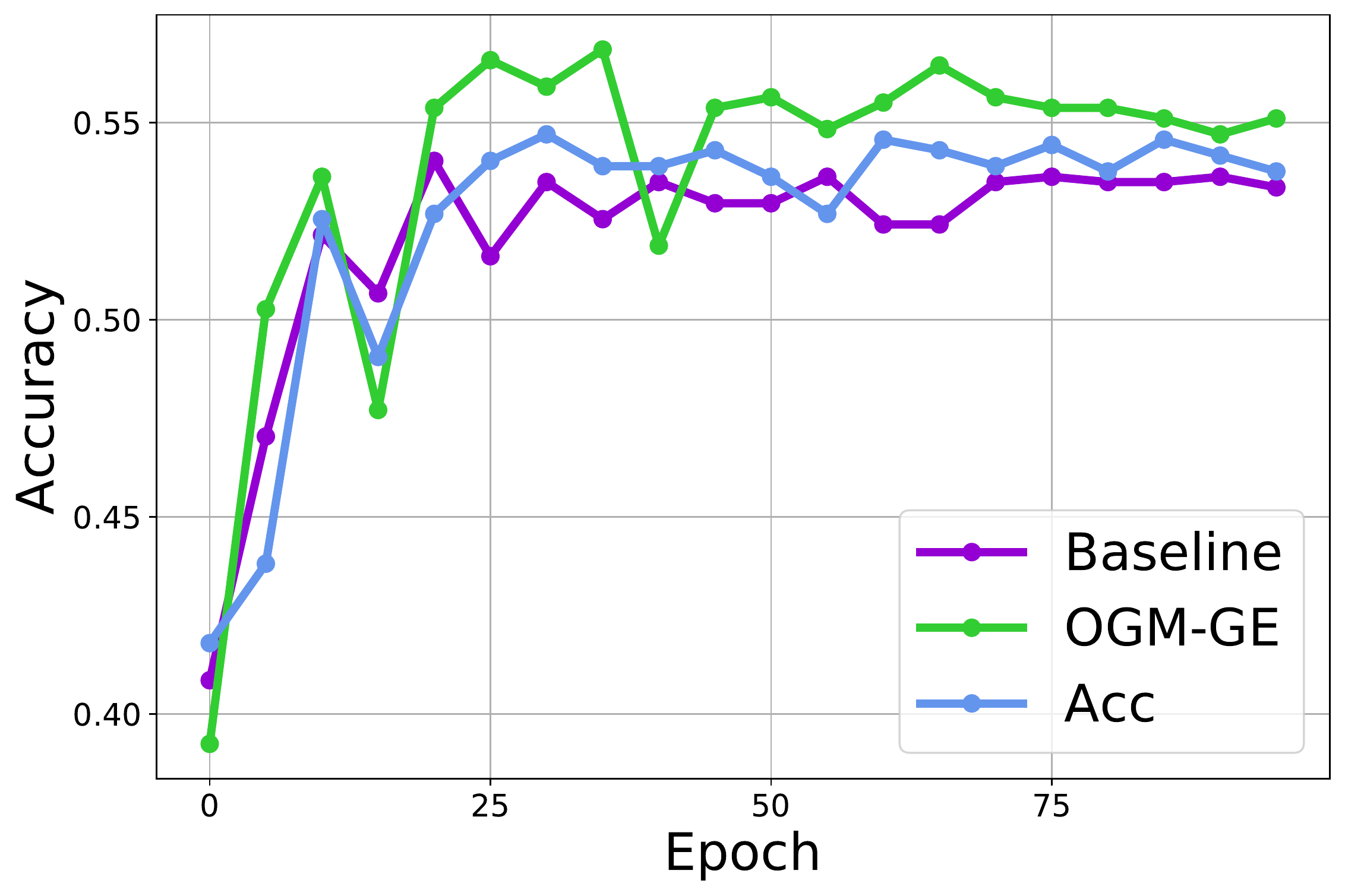}
    \caption{}
    \label{fig:performance}
  \end{subfigure}
  \hfill
  \begin{subfigure}{0.325\linewidth}
    %\fbox{\rule{0pt}{2in} \rule{.9\linewidth}{0pt}}
    \includegraphics[width=0.9\linewidth]{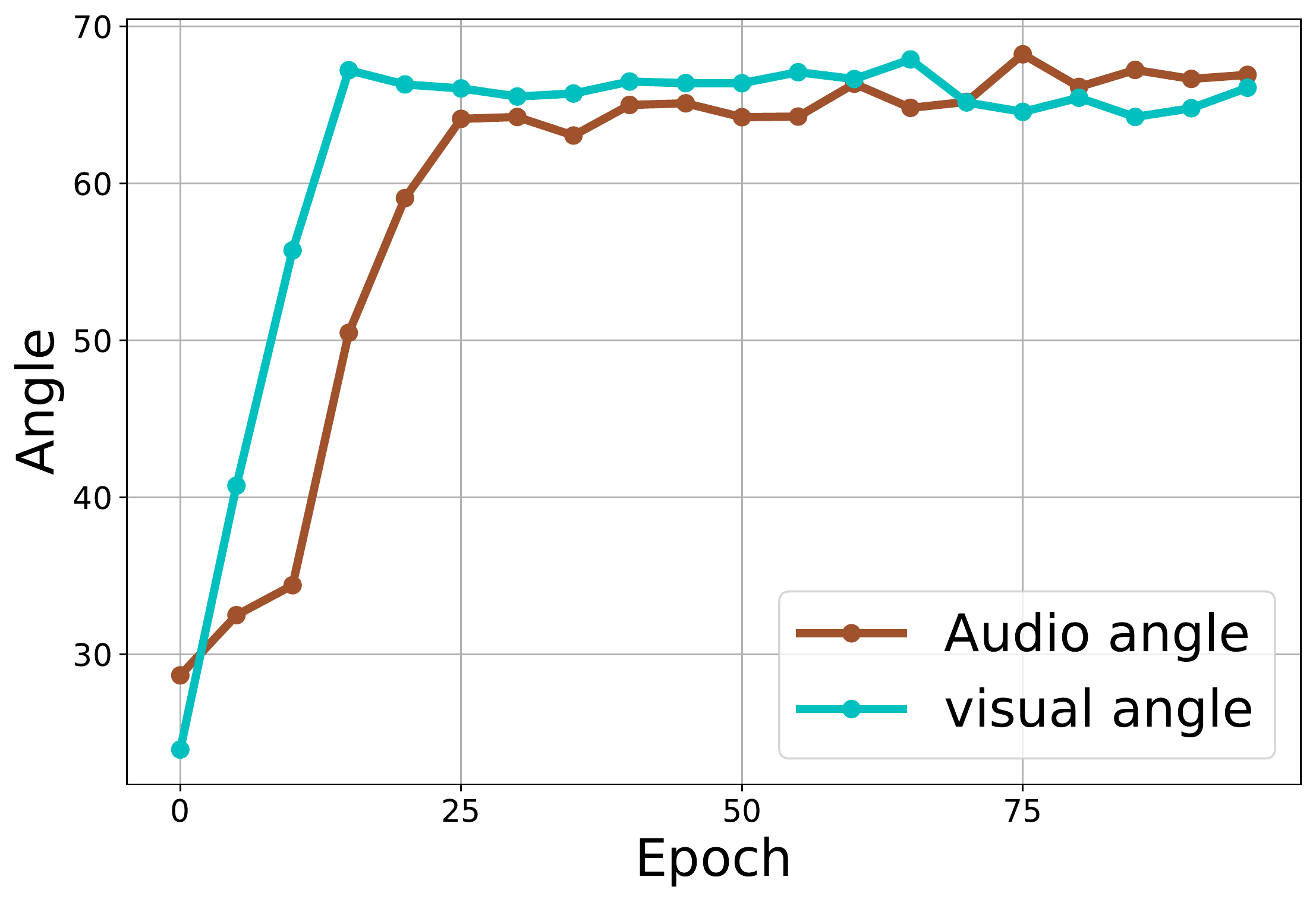}
    \caption{}
    \label{fig:angle}
  \end{subfigure}
  \caption{(a) Performance of each modality and their fusion. (b) Training accuracy of multimodal with different modulations. Baseline means no extra modulation. Acc increases the gradient magnitude of the slow-learning modality. OMG-GE \cite{peng2022balanced} reduces the gradient magnitude of better modality. (c) The gradient direction discrepancy (angle) between uni-modal and multimodal on the baseline. The results were acquired from CREMA-D.}
  \label{fig:Comparison and angle}
\end{figure*}

\subsection{Prototypical network}
Prototypical networks were originally proposed to solve few-shot or zero-shot classification problems~\cite{snell2017prototypical,roy2020few,ding2020graph,gao2019hybrid}, based on the idea that there is an embedding, defined as a prototype, which is surrounded by points from the same class.
Recently, this approach has been widely used to address long-tail recognition~\cite{yang2022multi}, domain adaptation~\cite{pan2019transferrable,chai2022dynamic}, and facilitate unsupervised learning~\cite{li2020prototypical}, since prototypes can be used to represent general features of a class.
In ~\cite{snell2017prototypical,li2020prototypical}, prototypes were interpreted as non-parametric prototypical classifiers that perform on par or even better than parametric linear classifiers. 
In this paper, we leverage prototypes to build non-parametric classifiers to evaluate the performance of each modality feature.

\section{Modality Imbalance Analysis}
\label{sec: Modality imbalance analysis}
Without loss of generality, we consider two input modalities as $m_0$ and $m_1$. The dataset is denoted as $\mathcal{D}$, which consists of instances and their corresponding labels $\left( \boldsymbol{x},y \right) $, where $\boldsymbol{x}=\left( \boldsymbol{x}^{m_0},\boldsymbol{x}^{m_1},y \right)=\left\{ x_{i}^{m_0},x_{i}^{m_1},y_i \right\} _{i=1,2,...,N}$, $y=\left\{ 1,2,...,M \right\} $, and $M$ is the number of categories . The goal is to train a model with this data to predict $y$ from $\boldsymbol{x}$.

We use a multimodal DNN with two uni-modal branches for prediction. Each branch has an encoder, denoted as $\phi ^0$, $\phi ^1$, to extract features of respective modal data $\boldsymbol{x}^{m_0}$ and $\boldsymbol{x}^{m_1}$. The representation outputs of encoders are denoted as $\boldsymbol{z}^0=\phi ^0\left( \theta ^0,\boldsymbol{x}^{m_0} \right) $ and $\boldsymbol{z}^1=\phi ^1\left( \theta ^1,\boldsymbol{x}^{m_1} \right) $, where $\theta ^0$ and $\theta ^1$ are the parameters of encoders. The two uni-modal encoders are connected through the representations by some kind of fusion, which is prevalent in multimodal learning~\cite{liu2018late,li2021align,zhang2022hierarchical}. In this work, we have tried some different fusion methods. For simplicity, we use $\left[ \cdot ,\cdot \right] $ to denote fusion operation. Let $W\in \mathbb{R} ^{M\times \left( d_{z^0}+d_{z^1} \right)}$ and $b\in \mathbb{R} ^M$ denote the parameters of the linear classifier to produce the logits output:
\begin{equation}
  f\left( \boldsymbol{x} \right) =W\left[ \phi ^0\left( \theta ^0,\boldsymbol{x}^{m_0} \right) ;\phi ^1\left( \theta ^1,\boldsymbol{x}^{m_1} \right) \right] +b
  \label{eq:f=wx+b}
\end{equation}

The cross-entropy loss of the multimodal model is 
\begin{equation}
    \mathcal{L} _{CE}=-\frac{1}{N}\sum\nolimits_{i=1}^N{\log \frac{e^{f\left( x_i \right) _{y_i}}}{\sum\nolimits_{k=1}^M{e^{f\left( x_i \right) _k}}}}
    \label{eq:CE loss}
\end{equation}

The gradient of the softmax logits output with true label $y^i$ should be:

\begin{equation}
  \frac{\partial \mathcal{L} _{CE}}{\partial f\left( x_i \right) _{y_i}}=\frac{e^{\left( W\left[ \phi ^0;\phi ^1 \right] +b \right) _{y_i}}}{\sum\nolimits_{k=1}^M{e^{\left( W\left[ \phi ^0;\phi ^1 \right] +b \right) _k}}}-1
  \label{eq:softmax gradient}
\end{equation}

For convenience, we simplify $\phi ^0\left( \theta ^0,\boldsymbol{x}^{m_0} \right)$ and $\phi ^1\left( \theta ^1,\boldsymbol{x}^{m_1} \right)$ as $\phi ^0$ and $\phi ^1$. According to \cref{eq:softmax gradient}, the final gradient is influenced by the performance of the fused modality. However, we cannot directly judge the contribution of the two modalities. To do it, we take a simple fusion method, summation, as the example here:
\begin{equation}
  \begin{split}
      f\left( \boldsymbol{x} \right) =W^0\cdot \phi ^0\left( \theta ^0,\boldsymbol{x}^{m_0} \right) +b^0\\
                                      +W^1\cdot \phi ^1\left( \theta ^1,\boldsymbol{x}^{m_1} \right) +b^1
  \end{split}
  \label{eq:sum fusion equation}
\end{equation}
where $W^0\in \mathbb{R} ^{M\times d_{z^0}}$, $W^1\in \mathbb{R} ^{M\times d_{z^1}}$ and $b^0,b^1\in \mathbb{R} ^M$ are the parameters for individual modal classifier. Therefore, we use the logits output of the ground truth as the performance for each modality and their summation fusion:
\begin{equation}
\begin{aligned}
  &s^0=\mathrm{softmax} \left( W^0\cdot \phi ^0+b^0 \right) _y \\
  &s^1=\mathrm{softmax} \left( W^1\cdot \phi ^1+b^1 \right) _y \\
  &s^{fu}=\mathrm{softmax} \left( W^0\cdot \phi ^0+b^0+W^1\cdot \phi ^1+b^1 \right) _y
  \label{eq:performance of each modality and sum}
\end{aligned}
\end{equation}

As shown in \cref{fig:performance comparison of each modality and their fusion}, the performance of the audio modal is very similar to the multimodal during training and the visual modal is much worse in CREMA-D~\cite{cao2014crema}, which means better modality contributes more to the gradient because of higher performance similarity with their fusion. Moreover, the visual modal is severely suppressed in the multimodal learning process because of the excessive dominance of gradient updates by the audio modal. Therefore, we have to mitigate the inhibition on visual modal to fully exploit visual features. A straightforward approach can be to increase the magnitude of the gradient. We test a similar way with OGM-GE~\cite{peng2022balanced}, named Acc, to increase the gradient magnitude of the slow-learning modality instead of lower the gradient magnitude of the better modality in OGM-GE. The results are shown in \cref{fig:performance,fig:angle}.  

\begin{figure*}[t]
  \centering
  \includegraphics[width=0.85\linewidth]{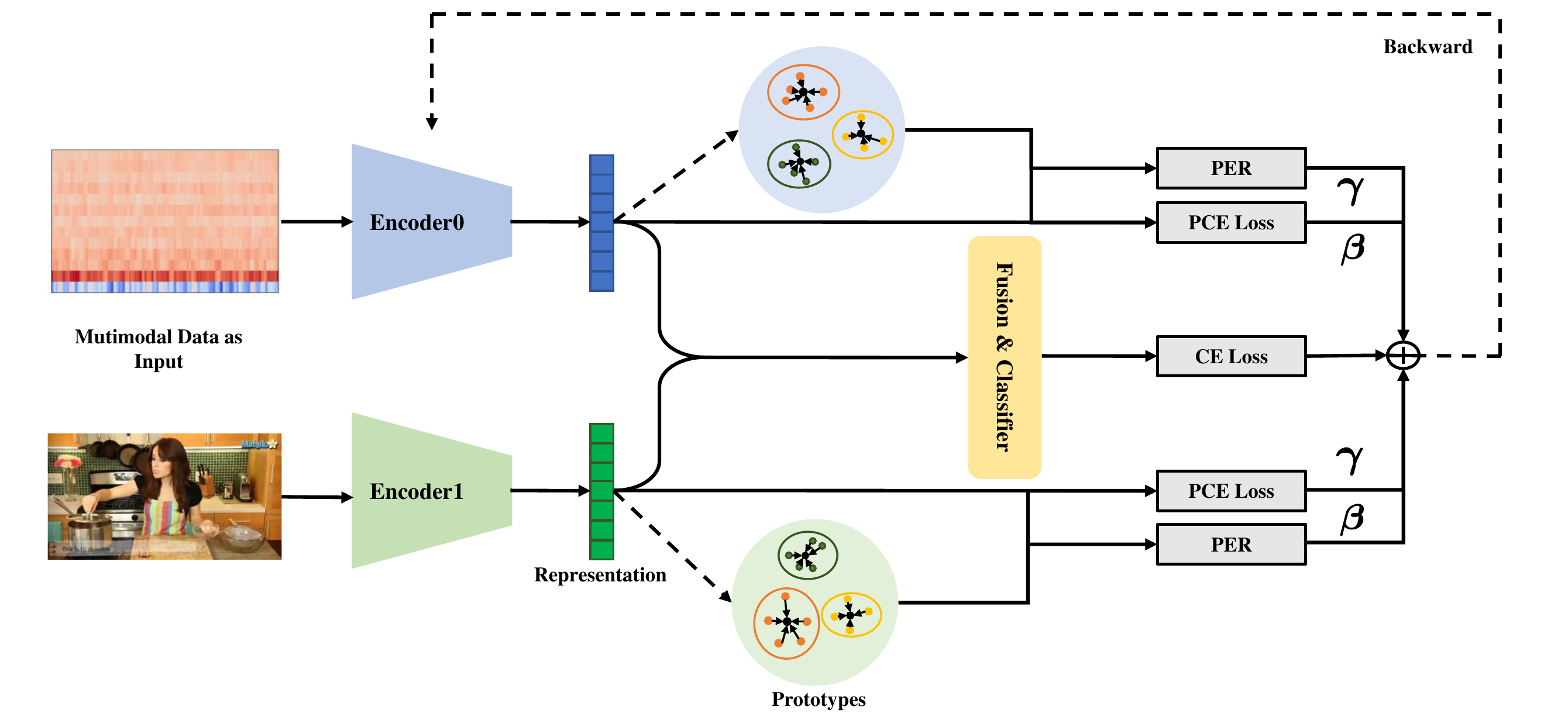}
  \caption{The pipeline of modality modulation with prototypical modal rebalance strategy.}
  \label{fig:pipeline of PMR}
\end{figure*}

As demonstrated in \cref{fig:performance}, increasing the gradient magnitude of the slow-learning modality (visual) could improve the validation accuracy a little bit but not as obviously as OGM-GE does. To find the reason for the phenomenon, we use the uni-modal output $s^0$ and $s^1$ to calculate the gradient for each modal branch additionally with CE loss and illustrate the direction discrepancy of gradients between each uni-modal and multimodal $s^{fu}$, as shown in \cref{fig:angle}. The angle between the actual gradient update direction (from the multimodal output) and each modal's guidance direction (from the uni-modal output) increases dramatically during training, in the meantime, remaining acute. Therefore, the two modalities influence each other, resulting in a larger gap between the gradient update direction obtained by the fused modality and the expected direction of each modality. That means the slow-learning modality cannot fully exploit its feature with the disturbance from other modalities, ultimately making the method of modulating the gradient amplitude limited, as illustrated in \cref{fig:prior and ours}.

\section{Prototypical Modal Rebalance}
\subsection{Prototypical CE loss for modal acceleration}
\label{sec: PCE loss for acceleration}
As discussed above, due to inconsistency in performance between modalities, they will affect each other on the exploitation of self-feature in multimodal learning, and the modality with worse performance is particularly suppressed, which limits multimodal performance. In order to solve the problem, we aim to facilitate the exploitation of the features of the slow-learning modality via prototypes, as shown in \cref{fig:pipeline of PMR}.

The performance estimation in \cref{eq:performance of each modality and sum} needs the logits output after the classifier and of course, the logits have to be decomposable into separate components of the two modalities. However, this constraint is too strong, limiting the estimation's application in most scenarios. Aiming to implement a universal estimation method, we introduce the prototypes of the categories. With the data $\boldsymbol{x}=\left\{ x_{i}^{m_0},x_{i}^{m_1},y_i \right\} _{i=1,2,...,N}$, we could produce the representations $\boldsymbol{z}=\left\{ z_{i}^{0},z_{i}^{1} \right\} _{i=1,2,...,N}$ in the training process. We denote $\boldsymbol{z}_{k}^{0}=\left\{ z_{k_i}^{0} \right\} _{i=1,2,...,N_k}$, $\boldsymbol{z}_{k}^{1}=\left\{ z_{k_i}^{1} \right\} _{i=1,2,...,N_k}$ as the subset data of each category, where $N_k$ is the number of class $k$, and $k=1,2,\cdots ,M$. The prototype is the centroid of each category data:
\begin{equation}
\begin{aligned}
  c_{k}^{0}=\frac{1}{N_k}\sum\nolimits_{i=1}^{N_k}{z_{k_i}^{0}} \\
  c_{k}^{1}=\frac{1}{N_k}\sum\nolimits_{i=1}^{N_k}{z_{k_i}^{1}} \\
  \label{eq:prototype}
\end{aligned}
\end{equation}

Then, we use prototypes to produce a distribution over classes for the input data $x$ based on a softmax over distances to the prototypes in the embedding space for each modality as in \cite{snell2017prototypical}:
\begin{equation}
\begin{aligned}
  p_{i}^{\phi ^0}\left( y=k|x_{i}^{m_0} \right) =\frac{\exp \left( -d\left( \phi ^0\left( x_{i}^{m_0} \right) ,c_{k}^{0} \right) \right)}{\sum\nolimits_{k'}^{}{\exp \left( -d\left( \phi ^0\left( x_{i}^{m_0} \right) ,c_{k'}^{0} \right) \right)}} \\
  p_{i}^{\phi ^1}\left( y=k|x_{i}^{m_1} \right) =\frac{\exp \left( -d\left( \phi ^1\left( x_{i}^{m_1} \right) ,c_{k}^{1} \right) \right)}{\sum\nolimits_{k'}^{}{\exp \left( -d\left( \phi ^1\left( x_{i}^{m_1} \right) ,c_{k'}^{1} \right) \right)}} \\
  \label{eq:prototype as classifier}
\end{aligned}
\end{equation}
where $d\left( \cdot ,\cdot \right)$ is the distance function, which is the Euclidean distance in this paper. Here, we design the imbalanced ratio inspired by \cite{peng2022balanced}:
\begin{equation}
    \rho _t=\frac{\sum\nolimits_{i\in B_{t}^{0}}^{}{p_{i}^{\phi ^0}}}{\sum\nolimits_{i\in B_{t}^{1}}^{}{p_{i}^{\phi ^1}}}
    \label{eq:imbalance ratio}
\end{equation}
$B_{t}^{0}$ and $B_{t}^{1}$ are the batch data at training step time $t$, therefore, we could use it to evaluate the degree of imbalance between two modalities in real-time. The metric is calculated using only representations and prototypes, which are both computationally independent of the fusion method and classifier structure. 

According to \cref{sec: Modality imbalance analysis}, in order to exploit more information of the slow-learning modality, increasing the gradient magnitude of the suppressed modal is not an ideal solution due to the perturbation by another modality. We leverage the prototypes to introduce the PCE loss that is independent of another modality to promote the slow-learning modal's performance:
\begin{equation}
  \begin{aligned}
    \mathcal{L} _{PCE}^{0}\left( f \right) =\mathbb{E} _{p\left( x_{i}^{m_0},y \right)}\left[ -\log \frac{\exp \left( -d\left( z_{i}^{0},c_{y}^{0} \right) \right)}{\sum\nolimits_{k=1}^M{\exp \left( -d\left( z_{i}^{0},c_{k}^{0} \right) \right)}} \right]  \\
    \mathcal{L} _{PCE}^{1}\left( f \right) =\mathbb{E} _{p\left( x_{i}^{m_1},y \right)}\left[ -\log \frac{\exp \left( -d\left( z_{i}^{1},c_{y}^{1} \right) \right)}{\sum\nolimits_{k=1}^M{\exp \left( -d\left( z_{i}^{1},c_{k}^{1} \right) \right)}} \right]  
    \label{eq:PCE loss}
  \end{aligned}
\end{equation}

The acceleration loss is the weighted combination of CE loss and PCE loss:
\begin{equation}
    \mathcal{L} _{\mathrm{acc}}=\mathcal{L} _{CE}+\alpha \cdot \beta \mathcal{L} _{PCE}^{0}+\alpha \cdot \gamma \mathcal{L} _{PCE}^{1}
    \label{eq:acc loss}
\end{equation}
where $\alpha$ is a hyper-parameter to control the degree of modulation. With $\rho _t$ to evaluate the imbalance degree dynamically, we are able to regulate the learning speed of each modality by adjusting the coefficients $\beta$ and $\gamma$ in a simple way:
\begin{equation}
  \left\{ \begin{matrix}
	\beta =clip\left( 0,\frac{1}{\rho _t}-1,1 \right) ,\gamma =0&		\rho _t<1\\
	\beta =0,\gamma =clip\left( 0,\rho _t-1,1 \right)&		\rho _t\geqslant 1\\
\end{matrix} \right. 
\label{eq:modulation method}
\end{equation}
where $clip\left( a,b,c \right)$ is the truncate function which constrains $b$ to be between $a$ and $c$. In this way, the slower-learning modality can be facilitated to exploit its features, while the better modality maintains the original learning strategy, which mitigates the modal imbalance phenomenon. Because the PCE loss only uses the representation of each modal encoder, our method can be applied to any modal fusion scenario, as long as the model itself has encoder(s) to extract features for two modalities respectively. Besides, to stabilize the learning process and reduce the computation cost, we compute the prototypes based on a subset of training data in a momentum fashion between each training epoch:
\begin{equation}
  \begin{aligned}
    c_{k}^{0}|_{old}=\varepsilon c_{k}^{0}|_{old}+\left( 1-\varepsilon \right) c_{k}^{0}|_{new} \\
    c_{k}^{1}|_{old}=\varepsilon c_{k}^{1}|_{old}+\left( 1-\varepsilon \right) c_{k}^{1}|_{new} 
    \label{eq:momentum prototype}
  \end{aligned}
\end{equation}
where $c_{k}|_{old}$ is the previous prototype in last epoch and $c_{k}|_{new}$ is the prototype calculated in the current epoch. 

\subsection{Prototypical entropy regularization for inhibition reduction}
With the help of PCE loss, we could accelerate the slow-learning modality. Nevertheless, the fusion output is still a hindrance that restrains the improvement of the slow-learning modality. As demonstrated in \cref{fig:performance comparison of each modality and their fusion,fig:angle}, the distraction from other modalities increases dramatically when the performance gap between modalities is quite distinct. In order to reduce the inhibition, we propose the prototypical entropy regularization (PER) terms to slow down the convergence speed of the dominant modality:

\begin{equation}
    \begin{aligned}
        \mathcal{L} _{final}=\mathcal{L} _{acc}-\mu \cdot \gamma H^0\left( \pi \left( -d\left( z^0,c_{y}^{0} \right) \right) \right) 
\\
\,\,               -\mu \cdot \beta H^1\left( \pi \left( -d\left( z^0,c_{y}^{1} \right) \right) \right) 
        \label{eq:prototypical entropy regularization}
    \end{aligned}
\end{equation}
where $\pi$ is the $softmax$ function which produces the probabilities and $H$ is the entropy. $\mu$ is a hyper-parameter. 
The coefficients $\beta$ and $\gamma$ are multiplied on the opposite modality compared with \cref{eq:acc loss}, which means accelerating slow-learning modality and preventing modality premature convergence happens at the same time. One point worth emphasizing is that we only add the regularization term in the first training few epochs to reduce the inhibitory effect in the early training stage to avoid performance damage of this modal. Overall, the pseudo-code of PMR is provided in \cref{alg:PAC}.

\begin{algorithm}[t]
\caption{multimodal learning with PMR.}
\label{alg:PAC}
\KwIn{Input data $\mathcal{D} =\left\{ x_{i}^{m_0},x_{i}^{m_1},y_i \right\} _{i=1,2,...,N}$, subset $\mathcal{D_s}$, initialized model parameters $\theta^0$, $\theta ^1$, hyper-parameters $\alpha$, $\varepsilon$, epoch number $E$, regularization epoch number $E_r$.}
\BlankLine
int e=0; \\
\While{\textnormal{$e < E$}}{
   Obtain the subset representations $ z_{s}^{0} $, $z_{s}^{1}$ by feeding-forward $\mathcal{D_s}$ to the model; \\
   Calculate the prototypes $c^0$, $c^1$ using \cref{eq:momentum prototype}; \\
   \ForEach{\textnormal{mini-batch data $B_t$ in $\mathcal{D}$ at step $t$}}{
      Obtain the representations $z_{t}^{0}$, $z_{t}^{1}$ by feeding-forward $B_t$ to the model; \\
      Calculate $\rho _t$ using \cref{eq:imbalance ratio}; \\
       \eIf{$e < E_r$}{
        Calculate the final loss $\mathcal{L} _{final}$ using \cref{eq:modulation method,eq:acc loss,eq:prototypical entropy regularization}; }
      {Calculate the final loss using \cref{eq:acc loss}}
      Update the model with the gradient based on $\mathcal{L} _{final}$;
   }
   e=e+1;
}
\end{algorithm}

\section{Evaluation}
\subsection{Datasets}
\textbf{CREMA-D}~\cite{cao2014crema} is an audio-visual dataset for the study of emotion recognition, which consists of facial and vocal emotional expressions. The emotional states can be divided into 6 categories: \textit{happy}, \textit{sad}, \textit{anger}, \textit{fear}, \textit{disgust} and \textit{neutral}. There are a total of 7442 clips in the dataset, which are randomly divided into 6698 samples as the training set and 744 samples as the testing set.

\textbf{AVE}~\cite{tian2018audio} is an audio-visual video dataset for audio-visual event localization, in which there are 28 event classes and 4,143 10-second videos with both auditory and visual tracks as well as second-level annotations. All the videos are collected from YouTube. In our experiments, we extract the frames from event-localized video segments and capture the audio clips within the same segments, constructing a labeled multimodal classification dataset. The training and validation split of the dataset follows~\cite{tian2018audio}.

\textbf{Colored-and-gray MNIST}~\cite{kim2019learning} is a synthetic dataset based on MNIST~\cite{lecun1998gradient}, which is denoted as CG-MNIST. Each instance contains two kinds of images, a gray-scale image, and a monochromatic image. In the training set, there are a total of 60,000 instances and the monochromatic images are strongly color-correlated with their digit label. In the validation set, the total number of instances is 10,000, while the monochromatic images are weakly color-correlated with its label. In this work, we consider the monochromatic image as the first modality and the gray-scale image as the second modality, the same setting in~\cite{wu2022characterizing}. We use this dataset to prove the method's effectiveness beyond the audio-visual dataset.

\subsection{Experimental settings}
For datasets CREMA-D and AVE, we adopt the ResNet18~\cite{he2016deep} as our encoder backbones and map the input data as 512-dimensional vectors. For audio modality, the data is converted to a spectrogram of size 257×1,004 for AVE and 257x299 for CREMA-D. For visual modality, we extract 10 frames from the video clips and randomly select 3 frames for CREMA-D (4 for AVE) to build the training dataset. For Colored-and-gray MNIST, we build a neural network with 4 convolution layers and 1 average pool layer as the encoder. We train all the models with mini-batch size 64 and an SGD optimizer~\cite{robbins1951stochastic} with a momentum of 0.9 and a weight decay of 1e-4. The learning rate is initialized as 1e-3 and gradually decays to 1e-4. The subset of data for prototype calculation is one-tenth the size of training data and is also extracted from the training data. $\alpha$ is set to 1 or 2, for different datasets. $\mu$ is set to a small value 1e-2 or 1e-3 for different datasets. All of our experiments were performed on one NVIDIA GeForce RTX 3090 GPU.  

\begin{table}
  \centering
  \begin{tabular}{c | c | c | c}
    \toprule
    \makebox[0.092\textwidth][c]{Dataset} & \makebox[0.092\textwidth][c]{CREMA-D} & \makebox[0.092\textwidth][c]{AVE} & \makebox[0.092\textwidth][c]{CG-MNIST} \\
    \hline
    Method & Acc & Acc & Acc \\
    \hline
    Uni-modal1 & 54.4 & 62.1 & 99.3 \\
    Uni-modal2 & 58.0 & 31.0 & 60.4 \\
    \hline
    Concatenation & 53.2 & 65.4 & 58.4 \\
    Summation & 52.9 & 65.1 & 59.1 \\
    Film & 57.2 & 64.4 & 60.0 \\
    Gated & 58.4 & 63.5 & 59.8 \\
    \hline
    Concatenation$\dagger $ & 61.1 & 67.1 & 77.2 \\
    Summation$\dagger $ & 59.4 & \textbf{68.1} & \textbf{78.5} \\
    Film$\dagger $ & \textbf{61.8} & 66.4 & 76.3 \\
    Gated$\dagger $ & 59.9 & 62.7 & 68.7 \\
    \bottomrule
  \end{tabular}
  \caption{Performance on CREMA-D, AVE and Colored-and-gray MNIST dataset with various fusion methods. $\dagger$ indicates PMR strategy is applied. PMR gets great performance improvement on nearly all scenarios}
  \label{tab:PAC compared with Normal}
\end{table}

\subsection{Effectiveness on the multimodal task}
\noindent\textbf{Comparison on conventional fusion methods.} In this experiment, we apply the PMR strategy on 4 kinds of basic fusion methods: concatenation \cite{owens2018audio}, summation, film~\cite{perez2018film} and gated~\cite{kiela2018efficient}. Among these, summation is the type of late fusion and the other three belong to intermediate fusion~\cite{iyengar2003discriminative,liu2013sample} method. The logit output of summation and concatenation can be split into two individual parts for each modality combined with a linear classifier, as discussed in \cite{peng2022balanced}. In film and gated, features between the modalities are fused in more complicated ways therefore the logit output cannot be fully split. The results are shown in \cref{tab:PAC compared with Normal}, which also includes the performance trained on every single modality. Modal1 is audio and modal2 is visual for CREMA-D and AVE, while modal1 is gray-modality and modal2 is colored-modality for CG-MNIST. According to the results, we can find that the performance of each uni-modal model in each dataset is inconsistent, as the audio performance is worse than the visual in CREMA-D, and on the contrary, the audio performance is better than the visual in AVE. In addition, uni-modal performance may outperform the vanilla fusion methods. For example, uni-visual performance is obviously better than the concatenation and summation for CREMA-D, so the uni-gray performance in CG-MNIST, indicating the inhibitory relationship between modalities. We get a significant improvement on the three datasets compared with each vanilla fusion method when our PMR strategy is applied, except for a slight decrease in gated on the AVE dataset. 
% This shows that our method is applicable and effective under multiple fusion methods.

\noindent\textbf{Improved performance compared with State-of-the-Art.} We compare our PMR strategy with other three modulation strategies for modality imbalance: Modality-Drop~\cite{xiao2020audiovisual}, Gradient-Blending~\cite{wang2020makes} and OGM-GE~\cite{peng2022balanced}. We compare them with concatenation and film fusion methods. We also demonstrate the performance of the baseline with the naive fusion method. All results are shown in \cref{tab:Comparison with SOTA}. We can notice that all the modulation methods achieve better performance compared with the baseline and our PMR strategy gets the best among them. The main improvement contribution comes from PCE loss to accelerate the slow-learning modality while combining with PER will still get almost a 1\% increase or at least stay the same. Apart from these, we also note that Gradient-Blending has to train additional uni-modal classifiers and requires the validation results with extra computation, while OGM-GE can only be directly used in concatenation but not film, therefore, we modulate the training process every few epochs as done in \cite{peng2022balanced}. Compared to them, our method doesn't need additional modules and is independent of the fusion methods and classifier structure, which makes it applicable to more scenarios. 

\begin{table}
  \centering
  \begin{tabular}{ c | c c | c c }
    \hline
    Dataset & \multicolumn{2}{c|}{CREMA-D} & \multicolumn{2}{c}{AVE} \\
    \hline
    Fusion & Concat & Film & Concat & Film \\
    \hline
    Naive & 53.2 & 57.2 & 65.4 & 64.4 \\
    Modality-drop & 55.8 & 58.3 & 66.4 & 63.9 \\
    Grad-Blending & 56.7 & 58.7 & 65.5 & 65.2 \\
    OGM-GE & 57.7 & 58.0 & 65.9 & 65.1 \\
    \hline
    PMR w/o PER & 60.3 & 61.3 & \textbf{67.2} & 65.6 \\
    \hline
    PMR & \textbf{61.1} & \textbf{61.8} & 67.1 & \textbf{66.4} \\
    \hline
  \end{tabular}
  \caption{Comparison with various modulation strategies on CREMA-D and AVE dataset with concatenation and film fusion methods. PMR achieves the best performance among these methods.}
  \label{tab:Comparison with SOTA}
\end{table}

\begin{figure*}[t]
    \centering
    \begin{subfigure}{0.325\linewidth}
        \includegraphics[width=0.9\linewidth]{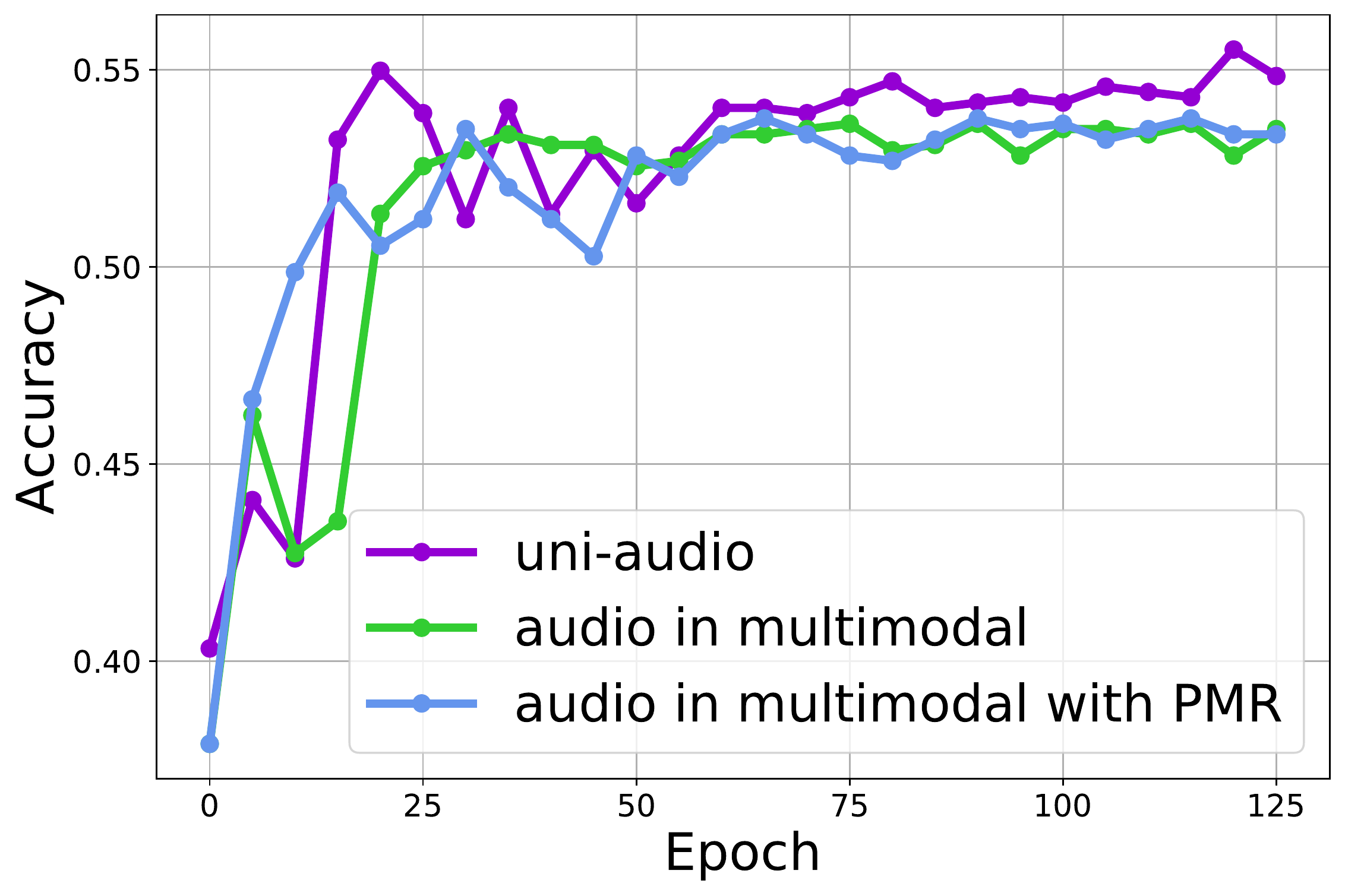}
        \caption{}
        \label{fig:uni-audio performance comparison}
    \end{subfigure}
    \hfill
    \begin{subfigure}{0.325\linewidth}
        \includegraphics[width=0.9\linewidth]{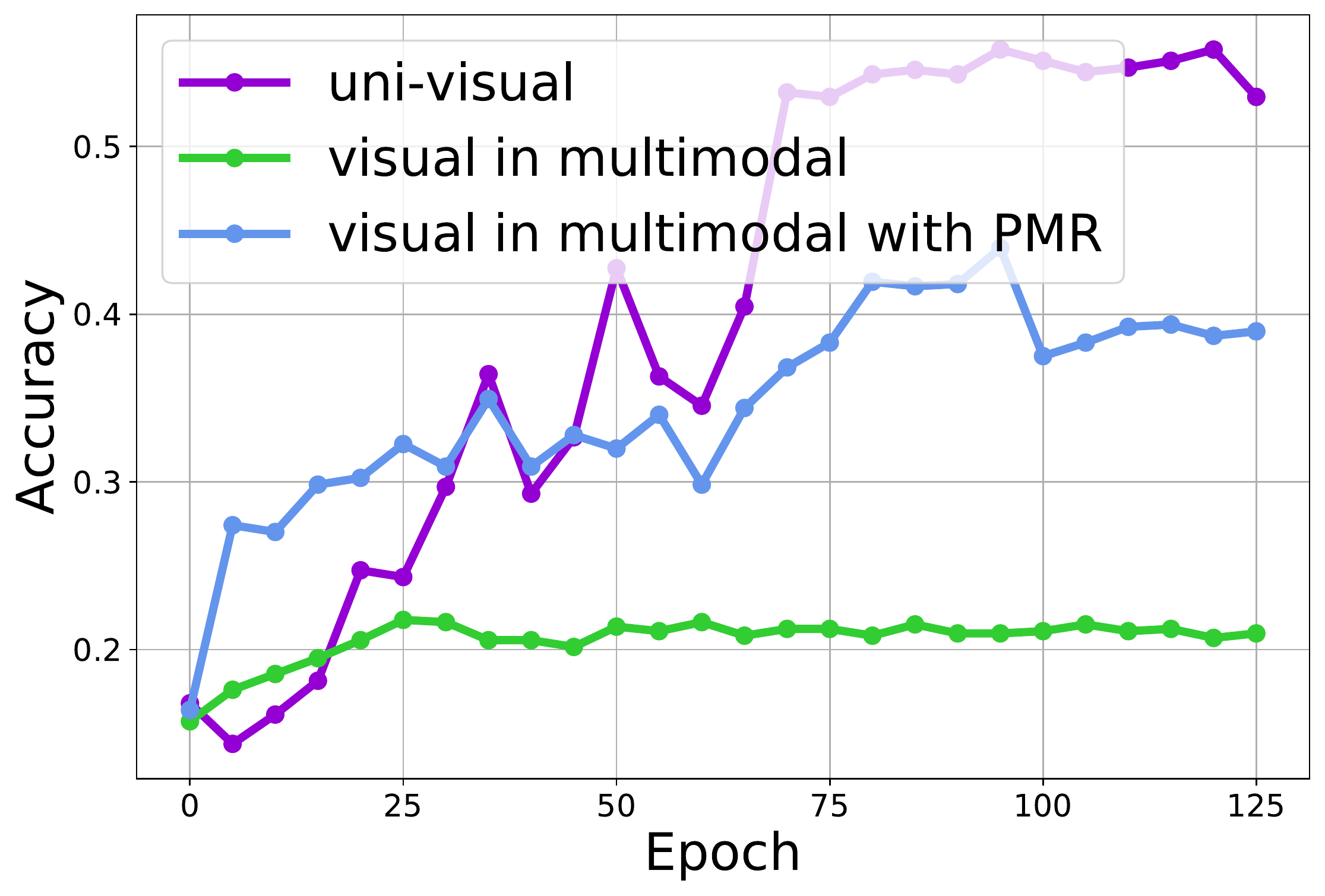}
        \caption{}
        \label{fig:uni-visual performance comparison}
    \end{subfigure}
    \hfill
    \begin{subfigure}{0.325\linewidth}
        \includegraphics[width=0.9\linewidth]{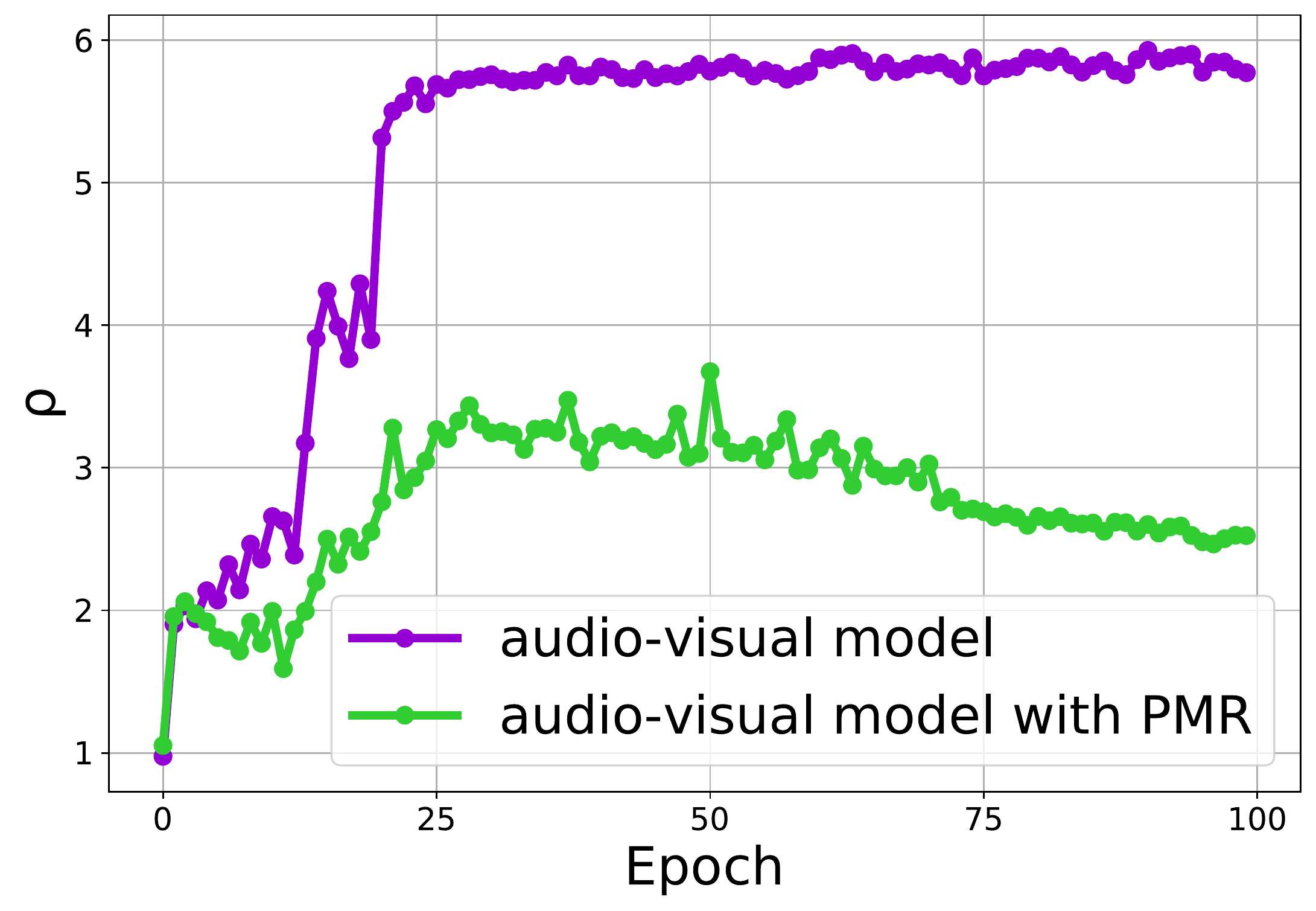}
        \caption{}
        \label{fig:imbalance ratio comparison}
    \end{subfigure}
    \caption{Performance of the uni-modal trained models, uni-modal branch in multimodal trained models, and uni-modal branch in multimodal with PMR on CREMA-D dataset. The fusion method in multimodal models is concatenation. (a) Performance of audio modality. (b) Performance of visual modality. (c) The change of imbalance ratio $\rho $.}
    \label{fig:uni-modal comparison}
\end{figure*}

\noindent\textbf{Performance on different architectures.}
The fusion stage of the four fusion methods used above is after the encoder or the classifier. To validate the applicability of PMR in more scenarios, we combine it with two intermediate fusion methods MMTM~\cite{joze2020mmtm} and CentralNet~\cite{vielzeuf2018centralnet} with and without PMR on CREMA-D and CG-MNIST. MMTM could recalibrate the channel-wise features of different CNN streams by squeeze and multimodal excitation steps. We use ResNet18 as the backbone and apply MMTM in the three final residual blocks. CentralNet uses both unimodal hidden representations and a central joint representation at each layer, which are fused by a learned weighted summation. We use only one frame for each video on CREMA-D for convenience. According to \cref{tab:comparison on intermediate fusion}, our proposed PMR achieves prominent improvement even if fusion is operated during the processing of encoders, indicating the applicability of PMR in complex scenarios. 

\begin{table}[t]
    \centering
    \begin{tabular}{c|c|c}
         \hline
         Dataset & CREMA-D & CG-MNIST \\
         \hline
         Method & Acc & Acc \\
         \hline
         MMTM & 51.6 & 71.4 \\
         CentralNet & 50.2 & 69.3 \\
         \hline
         MMTM$\dagger$ & \textbf{55.1} & \textbf{74.2} \\
         CentralNet$\dagger$ & 52.9 & 72.1 \\
         \hline
    \end{tabular}
    \caption{Performance on CREMA-D and CG-MNIST with two kinds of intermediate fusion methods. $\dagger$ indicates PMR strategy is applied, which achieves better performance.}
    \label{tab:comparison on intermediate fusion}
\end{table}

\noindent\textbf{Application on another task.} We process the AVE dataset as a classification dataset in the above experiments. Here we use the original AVE dataset to complete the audio-visual event localization task. We apply PMA on the AVEL~\cite{tian2018audio} with concatenation and DMRN~\cite{tian2018audio} fusion methods, which utilize LSTM to model temporal dependencies in the two modalities respectively. We evaluate the performance of supervised event localization in the late fusion style. In experiments, we only apply PMR on the ground truth video segments without extra processing on the video segments which are not relevant to the label. We choose to use Adam optimizer with the same settings in ~\cite{tian2018audio}. The results are shown in \cref{tab:event localization}. Applying PMR on the event localization task still achieves a certain promotion on accuracy. The utilization of two kinds of fusion methods on this task indicates that our method has great application potential in various task scenarios and various fusion methods.

\begin{table}[t]
    \centering
    \setlength{\tabcolsep}{7mm}
    \begin{tabular}{c|c}
         \hline
         Dataset & AVE \\
         \hline
         Fusion & Acc \\
         \hline
         concat & 72.7 \\
         DMRN & 73.1 \\
         \hline
         concat$\dagger$ & \textbf{74.3} \\
         DMRN$\dagger$ & 74.2 \\
         \hline
    \end{tabular}
    \caption{Event localization experiments based on the AVEL with two fusion methods. $\dagger$ indicates PMR strategy is applied. }
    \label{tab:event localization}
\end{table}

\subsection{Ablation study}

\noindent\textbf{Uni-modal performance comparison.} As we discussed in \cref{sec: Modality imbalance analysis}, there exists modality imbalance in multimodal learning. The interaction between modalities eventually leads to the failure of each modality to effectively exploit its own features. Therefore, we first compare the uni-modal performance among the uni-modal trained models and the uni-modal branch in the multimodal models with and without PMR strategy on CREMA-D dataset. As shown in \cref{fig:uni-audio performance comparison,fig:uni-visual performance comparison}, the uni-audio model is slightly better than the audio branch in vanilla multimodal learning, while the visual branch has few effective learning resulting in a distinct gap with uni-visual modal. After applying our proposed PMR strategy, the audio branch doesn't show a noticeable change since our strategy is mainly for facilitation for the slow-learning modality, while the visual branch improves considerably, indicating that our PMR strategy outperforms by indeed promoting the feature exploitation of the slow-learning modality. \cref{fig:imbalance ratio comparison} illustrates the change curves of imbalance ratio $\rho$ in vanilla multimodal and multimodal with PMR. It can be seen that our PMR helps to alleviate the modality imbalance phenomenon in multimodal learning. Although audio modality learns fast in the early training process and converges quickly, visual modality can still get improvement. Compare with the baseline without PMR, the imbalance ratio with PMR (green line) decreases gradually even if the dominant modality has already converged, indicating the intrinsic encouragement from PCE is less affected by other modalities.

\begin{table}
  \centering
  \begin{tabular}{ c | c | c }
    \hline
    \makebox[0.1\textwidth][c]{Dataset} & \makebox[0.1\textwidth][c]{CREMA-D} & \makebox[0.1\textwidth][c]{AVE} \\
    \hline
    Scale & Acc & Acc \\
    \hline
    baseline & 53.2 & 65.4 \\
    \hline
    1\% & 54.1 & 62.1 \\
    5\% & 58.4 & 65.8 \\
    10\% & 61.1 & 67.1 \\
    50\% & 61.2 & \textbf{67.7} \\
    100\% & \textbf{61.5} & 67.3 \\
    \hline
  \end{tabular}
  \caption{Experiments on CREMA-D and AVE with different subset scales. n\% indicates the proportion of the training dataset. Fusion method here is concatenation}
  \label{tab:subset scales}
\end{table}

\noindent\textbf{Analysis of subset data scales for prototype calculation.} In our proposed PMR method, the computation time of prototypes is the major additional cost. We analyze the performance of our method on different scales of data for prototype calculation. The results are shown in \cref{tab:subset scales}. When the scale of subset data is greater than 10\%, PMR tends to be stable in accuracy, which means data of this scale can represent the approximate distribution of the overall data on CREMA-D and AVE. If the data scale is too small, the performance of the model would drop a little bit, even worse than the vanilla baseline when the subset data scale is just 1\% on AVE. The reason may be the calculated prototype is biased because of the limited data size, which further hinders the training of the model. Although our method introduces a certain amount of extra computation, the required computation is not much and can be reduced with a reasonable selection of data.

\noindent\textbf{Adaptive optimizers.} We analyze the modality imbalance problem and propose the PMR strategy under the assumption: the same learning rate for each gradient calculation, which is true for popular SGD-based algorithms, but not rigorous for adaptive optimization algorithms. To validate the effectiveness of our method on more adaptive methods, we apply PMA on optimizers AdaGrad~\cite{duchi2011adaptive} and Adam~\cite{kingma2014adam}, dynamically adjusting the learning rate for each parameter. As demonstrated in \cref{tab:different optimizers}, we empirically show that PMA works with different optimizers. Different optimizers perform inconsistently on different datasets, e.g. Adam achieves the best performance on CREMA-D and SGD is the greatest in AVE.

\begin{table}[h]
    \centering
    \setlength{\tabcolsep}{5mm}
    \begin{tabular}{c|c|c}
         \hline
         Dataset & CREMA-D & AVE \\
         \hline
         Optimizer & Acc & Acc \\
         \hline
         SGD & 53.2 & 65.4 \\
         AdaGrad & 58.8 & 65.1 \\
         Adam & 59.8 & 64.4 \\
         \hline
         SGD$\dagger$ & 61.1 & \textbf{67.1} \\
         AdaGrad$\dagger$ & 61.5 & 66.2 \\
         Adam$\dagger$ & \textbf{65.3} & 66.5 \\
         \hline
    \end{tabular}
    \caption{Experiments with AdaGrad and Adam optimizers on CREMA-D and AVE. $\dagger$ indicates PMR strategy is applied. PMR consistently achieves better performance.}
    \label{tab:different optimizers}
\end{table}

\section{Discussion}
Multimodal learning usually falls into a suboptimal solution because of the modality imbalance problem, which indicates that vanilla optimization with a single strategy for different modalities is limited. We propose the prototypical modal rebalance (PMR) strategy to introduce different learning strategies for different modalities, i.e., accelerating the slow modality with prototypical cross entropy (PCE) loss and reducing the inhibition from dominant modality with prototypical entropy regularization (PER) term. This method achieves considerable performance improvement on the three multimodal datasets with different model structures and fusion methods. The non-parametric classifiers with prototypes can be applied in any scenario as long as we have the representations of instances for each modality.

%%%%%%%%% REFERENCES
{\small
\bibliographystyle{ieee_fullname}
\bibliography{egbib}
}

\end{document}